\documentclass[10pt, a4paper]{article}
\usepackage{lrec2022} % this is the new LREC2022 Style
\usepackage{multibib}
\newcites{languageresource}{Language Resources}
\usepackage{graphicx}
\usepackage{tabularx}
\usepackage{soul}
\usepackage{subcaption}
\usepackage{amsmath}
\usepackage{amsfonts}
\usepackage{booktabs}
\usepackage{microtype}
\usepackage{pifont}
\usepackage{titlesec}
\titleformat{\section}{\normalfont\large\bfseries\center}{\thesection.}{1em}{}
\titleformat{\subsection}{\normalfont\SmallTitleFont\bfseries\raggedright}{\thesubsection.}{1em}{}
\titleformat{\subsubsection}{\normalfont\normalsize\bfseries\raggedright}{\thesubsubsection.}{1em}{}
\renewcommand\thesection{\arabic{section}}
\renewcommand\thesubsection{\thesection.\arabic{subsection}}
\renewcommand\thesubsubsection{\thesubsection.\arabic{subsubsection}}
\usepackage{epstopdf}
\usepackage[utf8]{inputenc}

\usepackage[hidelinks]{hyperref}
\usepackage{xstring}

\usepackage{color}

\title{BERTHA: Video Captioning Evaluation Via Transfer-Learned Human Assessment}

\name{Luis Lebron$^{1}$, Yvette Graham$^{2}$, Kevin McGuinness$^{1}$, Konstantinos Kouramas$^{3}$, \\ {\bf \large  Noel E. O'Connor$^{1}$ }} 

\address{$^{1}$Insight SFI Research Centre for Data Analytics, Dublin City University (DCU)\\ $^{2}$School of Computer Science and Statistics, Trinity College Dublin\\ $^{3}$Collins Aerospace \\
         luis.lebroncasas@insight-centre.org\\}

\abstract{
Evaluating video captioning systems is a challenging task with multiple challenges to consider. Firstly, the fluency of the caption, multiple actions taking place within a single scene, and estimation of what a human user might consider important in a video. Most metrics aim to measure how similar the system generated captions are to a single or a set of human-generated captions. This paper presents a new method based on a deep learning model to evaluate systems. The model is based on BERT language model, shown to work well across a range of NLP tasks. The aim is for the model to learn to perform an evaluation similar to that of a human. To do so, we use a dataset that contains human evaluation of system-generated captions. The dataset consists of human judgments of the quality of captions produced by the system participating in past TRECVid video to text tasks \cite{1178722}. These annotations will be made publicly available.\footnotemark The new video captioning evaluation metric, BERTHA, obtains favourable results, outperforming commonly applied metrics in some setups.
 \\ \newline  }

\begin{document}
\maketitleabstract 
\footnotetext{https://github.com/LLebronC/TRECvid\-VTT\_HA}
\section{Introduction}
\label{intro}

Automatic video captioning is a challenging multimodal task requiring the successful combination of computer vision and natural language generation. Systems aim to generate fluent natural language for videos of various duration, ranging from generation of individual sentences that describe short videos \cite{awad2020trecvid} to several sentences for longer videos \cite{tang2019coin,caba2015activitynet}. 

This task has many challenges, with evaluation of the task itself being one of the most problematic. Video captioning differs from other tasks by looking for accurate descriptions of specific objects and also requires accurate  description of the context. A significant challenge in evaluating video captioning systems lies in the fact that there legitimately exists multiple and often very many distinct ways to describe what took place in a single video.

\begin{figure}[t]
    \centering
    \includegraphics[width=.9\linewidth]{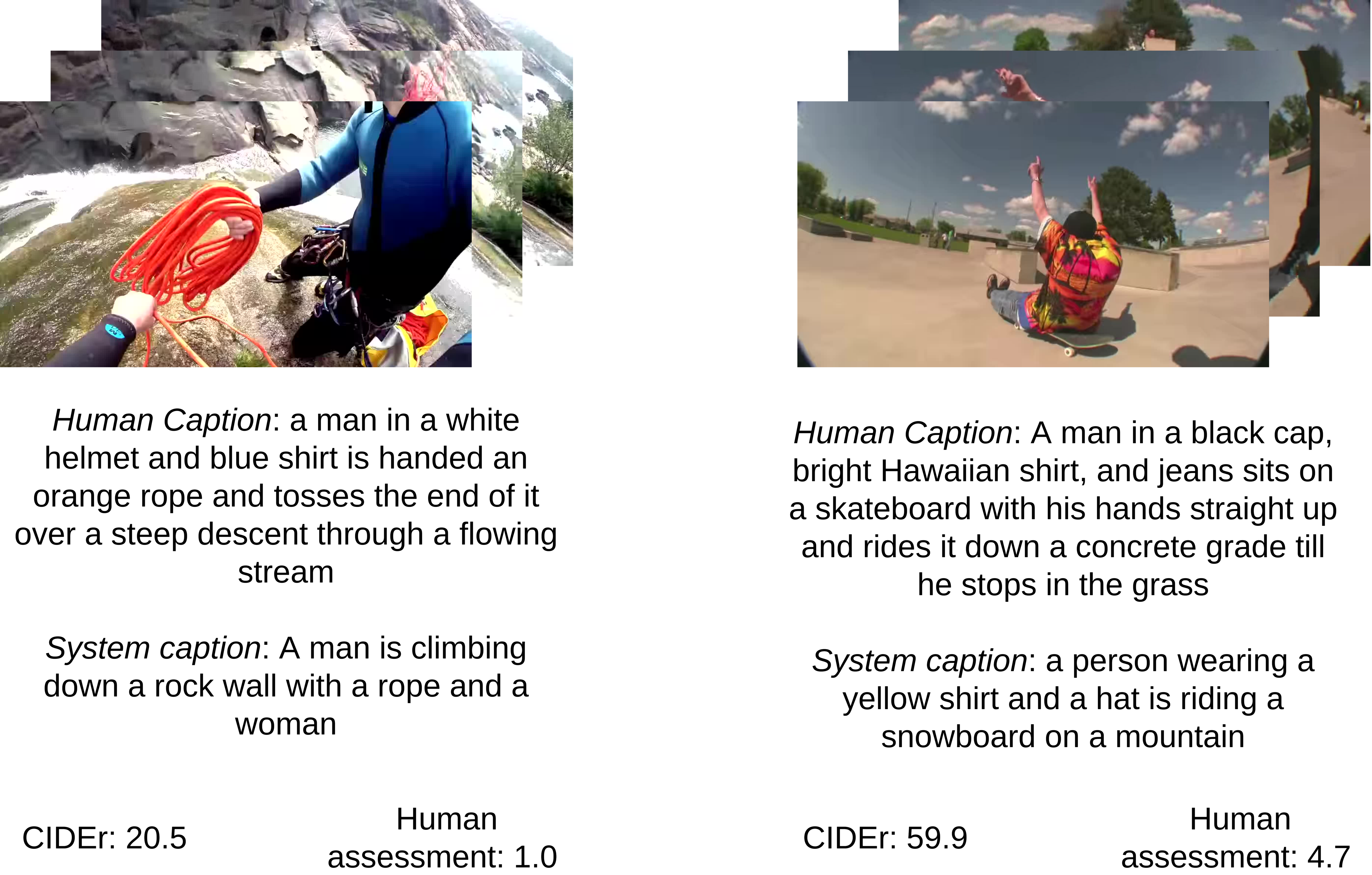}
    \caption{In these two pictures, we can see an example of captions where a popular automatic metric (CIDEr) fails to accurately evaluate the caption. In the first case, the system caption does not contain as much detail as the human caption, and it includes a woman who is nowhere in the scene. In the second case, the system caption seems like it does not describe the same scene; however, it contains some keywords in the human caption.}
    \label{fig:examples_auto_metric}
\end{figure}

The usual automatic metrics used for video captioning are borrowed from other tasks like machine translation or image captioning. Metrics like BLEU \cite{papineni2002bleu}, METEOR  \cite{banerjee2005meteor}, or CIDEr \cite{vedantam2015cider} are normally based on computing the overlap between the caption and the reference sentence. However, they fail to include the context of the scene. %They fail to correlate with human judgment as they do not consider the context. 
New metrics like SPICE \cite{anderson2016spice} try to address this problem using a graph to represent the semantic content.  

Evaluating the accuracy of a given metric is ordinarily carried out by computing the correlation of its scores for systems with human assessment of the quality of system-generated captions. 
%In other words, examining the degree to which metric scores for captions correspond with the human opinion of the same caption. 
In this paper, we present a new automatic metric that directly learns from human evaluation to maximize the correlation with human judgement. Our new method of evaluating video captioning systems makes use of pre-trained language models, BERT \cite{devlin2018bert}, to learn this correlation. 
%In general terms, language models aim to predict the next word or character within a text. Recent work has seen substantial advances made by the application of models trained on vast quantities of data to language processing tasks and the emergence of well-known language models such as  and GPT-3~\cite{brown2020language}.

By fine-tuning a pre-trained language model, it is possible to obtain excellent results in a huge variety of tasks \cite{devlin2018bert}, which overall is less computationally demanding than training a model from scratch. 
BERT has been shown to work well for tasks similar to video captioning such as MT evaluation \cite{sellam2020bleurt}, and thus we adopt this model in our work.

%Since existing metrics do not correspond well with human judgment about what may or may not be a good quality caption~\ref{fig:examples_auto_metric}, our proposal is a new metric based on BERT and fine-tuned with an accurate human evaluation of video captions. 
We test the performance of our metric on data from past TRECVid \cite{1178722} video to text challenge tasks. TRECVid is a well-known challenge where one of the tasks, the video to text task (VTT), is to produce sentences for short clips. The data used contains human judgments for the captions created by the participants' systems. 
As human judgment is sometimes unreliable, we collect additional multiple human assessments per caption to reduce the uncertainty contained within the training and test data. 

The contributions presented in this paper are as follows: 
\begin{itemize}
\item A new video captioning metric trained in human judgments that employs a pre-trained language model to aide the training phase.
\item An evaluation of the performance of the metric and how it compares with other commonly used metrics in a challenging dataset. The new metric is consistently in the top-performing metrics.
\item A study of the behavior of the metric under different scenarios and a test of the limits of the metric to further understand its performance. 
\end{itemize}

The remainder of the paper is structured as follows: we begin with a brief description of related work in Section \ref{related_work}. Next in Section \ref{method}, the model is described. Section \ref{results} contains the evaluation of the proposed metric in TRECvid VTT is analysed. The results show good performance compared with commonly used metrics. The paper finishes in Section \ref{conclusions} with some conclusions.

\section{Related Work}
\label{related_work}

\subsection{Video Captioning Metrics}

%The evaluation of captions can be divided into two factors: the quality of the generated text and how well it describes the scene. For the first, there are many metrics in the NLP domain that can evaluate this to a certain point. For instance, language models like BERT~\cite{devlin2018bert} or GPT ~\cite{radford2018improving}, can measure the probability that a set of words can occur in a sentence. Also, part-of-speech-tagging or parsing ~\cite{manning2014stanford} can be used to find the category of the word, e.g. noun, verb, etc., and identify mistakes. Finally, sentiment analysis can evaluate sentences more subjectively and identify which sentiment the phrase is transmitting to the reader~\cite{devlin2018bert}. However, the second factor is substantially more challenging as is more subjective to the human reading the caption.

Different automatic evaluation techniques use various aspects to measure the generated captions, and usually people employ multiple metrics as they can complement each other. In most evaluations, the main points to consider are fluency of the generated caption, accuracy in describing the content of the video, and similarity to the human references, with the latter being the basis of most common metrics.

One of the first metrics to be employed for video captioning was \textit{subject verb object (SVO)} accuracy. However, this metric is no longer typical in evaluation of video captioning systems as it is limited to measuring accuracy based on small set of words. BLEU \cite{papineni2002bleu}, which was initially developed for machine translation, is a metric based on computing the geometric mean of $n$-gram match counts.% As with the SVO metric, it does not account for the relationship between words in the language.
ROUGE \cite{lin2004rouge} is similar to BLEU, which also uses $n$-grams. The difference is that ROUGE considers the $n$-gram occurrences in the total sum of the number of reference sentences while BLEU considers the matches in the sum of candidates. %All these metrics only look for exact tokens, which is a big disadvantage when rephrasing.

In some cases, we not  only have a single reference per video but a set of them. A metric like METEOR \cite{banerjee2005meteor} performs better in these situations. It compares exact token matches, stemmed tokens,  paraphrase matches, and semantically similar matches using WordNet synonyms. 

Another aspect to consider is differences across references. Metrics like CIDEr \cite{vedantam2015cider} try to take this into account by measuring the consensus between the set of reference sentences. Another approach is SPICE \cite{anderson2016spice}, which is a metric that uses a scene graph to describe the semantics allowing it to identify similar sentences in a more generic way. 

\subsection{ Human Judgment}

Scores produced by such metrics aim to evaluate the similarity between a system-generated caption and a human reference caption. However, they cannot detect if they are describing the same scene. For instance, in a clip, there can be multiple details that are not important for the human annotator, like the colour of an animal, but the model can decide to include it in the final caption. An example of this problems can be seen in Figure~\ref{fig:examples_auto_metric}. Most metrics additionally have problems in terms of robustness, systems being penalized for legitimate differences in word order, using word replacement or change in the given score when using a shorter version of the original sentence. As this becomes difficult to compare automatically, a human judgment is normally involved to compare the metrics. This requires direct human evaluation of the caption quality or human-generated captions to use as a reference.
%When a human-generated reference caption is employed, a quantitative evaluation of sorts can be carried out that measures how similar the system-generated caption is to the human reference.
Then metrics can be compared based on how well they correlate to this human judgment.

There have been some studies on training models in the machine translation domain to imitate human evaluations of the sentences with actual human scores annotators. In this use case, most methods aim to solve this problem with handcrafted features and classical machine learning methods. However, more recently, participants of the WMT Metrics Shared task \cite{mathur2020results} have begun to include deep learning, specifically BERT as a base architecture \cite{sellam2020bleurt,shimanaka2019machine} and have obtained good results. 

Similar to the machine translation domain, some studies are looking for a way to learn a metric from human judgments. Datasets like Microsoft COCO~\cite{chen2015microsoft}, or PASCAL~\cite{vedantam2015cider}, have some captions annotated and judged by humans. In the case of COCO, the captions are evaluated with these criteria: percentage of captions that are better or equal to humans; percentage of captions that pass the Turing Test; average correctness of the captions; the average amount of details; and percentage of captions that are similar to human references. Most of the other datasets only use a scoring from 1 to 5,or similar, to evaluate how similar a caption is to a human reference or how well it describes the clip. Even the more complex evaluation done in COCO falls short of establishing a satisfactory ranking between all the captions as the number of evaluation and systems are very limited. The study of trained metrics~\cite{yi2020improving,cui2018learning} in these datasets has demonstrated that they correlate better with human judgment than the typical image captioning metrics.

\subsection{Language Models and Transfer Learning}

Language models are commonly employed in a wide range of NLP tasks and hugely influence text fluency produced by automated systems. They are models which predict the probability that a sequence of words can occur in a sentence.  Most architectures involve training on huge unlabelled corpora such as Wikipedia and news articles, literary text, and language-based web data~\cite{brown2020language}. This type of large architecture aims to decouple the model from the task so that the same model can be applied to distinct tasks. 

One example of this model is BERT \cite{devlin2018bert}, which is a bidirectional language model based on Transformers \cite{vaswani2017attention}. BERT has become very popular thanks to performing well in a diverse set of tasks like semantic textual similarity or sentiment analysis. This is achieved using a technique call transfer learning. These techniques consist of using an already trained model to solve another task by reusing the knowledge already learned in the new task. In BERT, the base model is trained on unlabeled textual data using two techniques: masked language modelling and next sentence prediction. Then this pre-trained model can be fine-tuned on multiple tasks. This approach has been demonstrated to work well in multiple similar tasks like machine translation evaluation \cite{sellam2020bleurt,shimanaka2019machine} or image captioning evaluation~ \cite{cui2018learning}. We follow the same paradigm and use BERT as the base model for our experiments.

\begin{figure*}[!ht]
\centering
\begin{subfigure}{.5\textwidth}
  \centering
  \includegraphics[width=.9\linewidth]{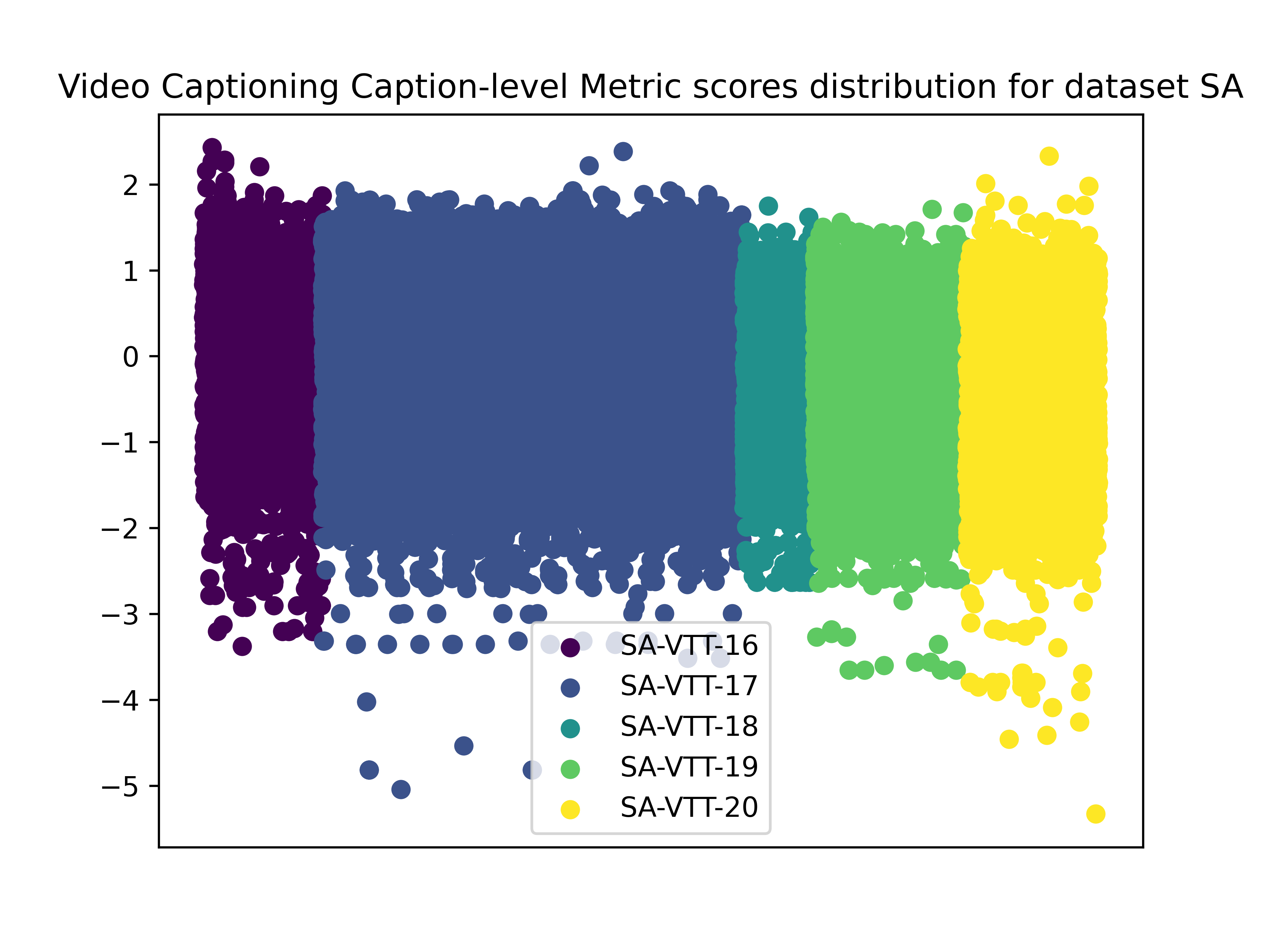}
  \caption{Score distribution of video captioning caption-level metric scores with human assessment SA for TRECvid 2016 to 2020 }
  \label{fig:sub1}
\end{subfigure}%
\begin{subfigure}{.5\textwidth}
  \centering
  \includegraphics[width=.9\linewidth]{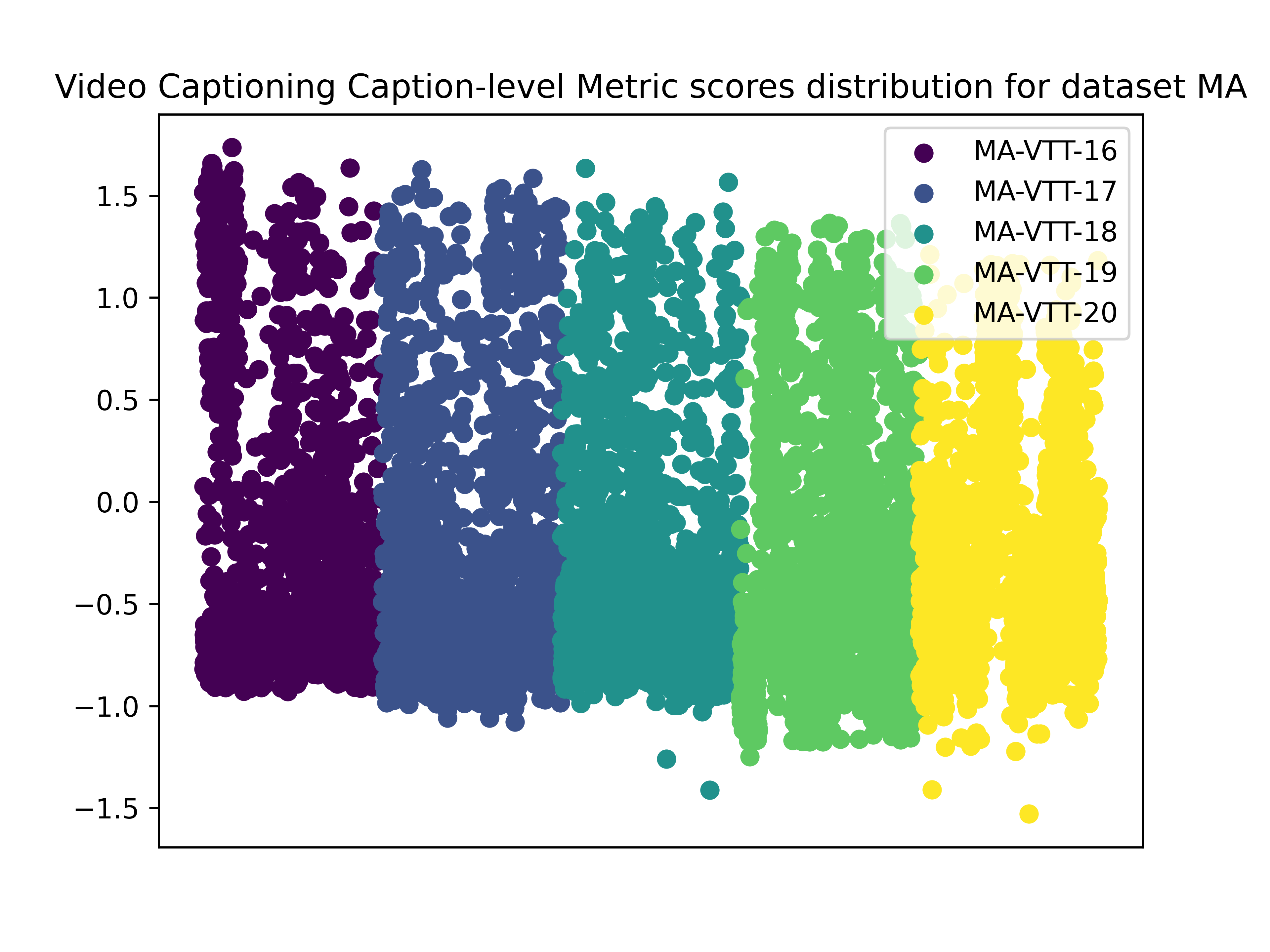}
  \caption{Score distribution of video captioning caption-level metric scores with human assessment MA for TRECvid 2016 to 2020 }
  \label{fig:sub3}
\end{subfigure}
  \caption{Score distribution of video captioning caption-level metric scores with human assessment for both datasets }
  \label{fig:captions}
\end{figure*}

\section{Evaluation of Video Captioning via Transfer Learning}
\label{method}
%\begin{figure}
%    \centering
%    \includegraphics[width=.9\linewidth]{figs/BERTHA_diagram_v2-5.png}
%    \caption{Evaluation with proposed BERTHA video captioning evaluation, where the input to the model is a system-generated caption and a human reference caption. Video Captioning System %and Human Caption Annotator correspond  to  one of the systems $M$ in the dataset and one of the human references use $S$, with $m \in M$ and $s \in S$. The bottom part refers to the human assessment used to obtain the human score $A$. Two symbols are added to the captions: one to mark the beginning of the sentences [CLS] and another to mark the end of the individual phrase [SEP]. The first output of the model, denoted by the symbol $C$, is feed to a Fully Connected layer (FC) with Softmax activation to obtain the predicted score $\gamma$. }
%    \label{fig:bertha_diagram}
%\end{figure}
%A good caption generation system can produce similar captions to those written by humans. 
When automating the evaluation of tasks aimed at human users, such as video captioning, a common approach is to compare system outputs with one or more human-generated reference(s). In the proposed model, we only use a single pair of caption and human references each time. This simple scenario is helpful as it can work with multiple human-generated references or with a limited number of them. 

Given a system output caption and a human-generated reference, our proposed evaluation model aims to learn the distance in meaning between the two strings of text. To train the model, we use captions output by systems participating in past benchmark shared tasks and carry out the human evaluation of the quality of the captions using Direct Assessment \cite{grahametal:2015,graham-etal-2020-assessing}. The details of the data collection process are explained in the next section.

%Human evaluation takes the form of a crowd-sourced worker watching the video, reading the caption and rating how well the caption describes what took place in the video.
%these distances from human-annotated captions. 
%Thereafter, a set of system-generated captions are assessed by humans who watch the video in question and provide a rating that reflects the quality of the system-generated caption.
%The system caption, a corresponding human-generated reference caption (also taken from data employed in benchmark tasks) and the human assessment score we collect are then employed to train the model, with the
%In these annotations, a score was given depending on how similar the description was to a references sentences. 
%aim of the model to maximize the correlation between the learned metric and the human ratings of caption quality.

%\subsection{The input to model}
We define $V=\{v_1,...,v_N\}$ as the set of $N \in \mathbb{N}$ videos. Notice that we can have $M \in \mathbb{N}$ different reference captions for a single video.
%with $N \in \mathbb{N}$ the number of videos, each video then can have up to $M \in \mathbb{N}$ captions $C = \{c_{1,1},...,c_{N,1},...,c_{N,M}\} $. 
We employ $S \in \mathbb{N}$ captioning systems to produce captions for videos and carry out $\mathrm{A}_{N',M',S} \in \{x \in \mathbb{R} \mid 0 \leq x \leq 1\} $ human evaluation of video captioning systems where $N' \leq N $ and $M' \leq M $. Note that not all the videos in the source dataset are annotated. 
The goal of training the model is to maximize the correlation, $\rho(\Gamma_{\theta},\mathrm{A})$, between predicted scores ($\Gamma_{\theta} \in \mathbb{R}^{N'\times M'\times S}$) and the given human qualitative scores $\mathrm{A}$:
\begin{equation}
    \operatorname*{argmax}_\theta\; \rho(\Gamma_{\theta},\mathrm{A}),
\end{equation}
where $\rho$ is the Pearson product-moment correlation coefficient.
%
%
%\subsection{BERTHA}
Even if we used all the data available for the task, it is a relatively small dataset. We therefore use a pre-trained language model and fine-tune it for the task.
%Figure \ref{fig:bertha_diagram} provides a depiction of the process involved in training our proposed video captioning evaluation metric, BERTHA. The model is based on BERT (Bidirectional Encoder Representations from Transformers)~\cite{devlin2018bert}.
The core functionality of BERT is a set of transform blocks \cite{vaswani2017attention}, which can have two sentences as input, and output two sets of symbols corresponding to these input sentences.

BERTHA use the BERT-base as it is the smaller model of the two presented in the original paper. It consists of 12 layers, 768 hidden states, and 12 heads. The BERT-base is trained with the configuration described in  \cite{devlin2018bert}. The base model is pre-trained using two unsupervised tasks:  masked language modelling,  and subsequent sequence prediction. The dataset used is BooksCorpus \cite{zhu2015aligning} and English Wikipedia. Using this model as a baseline, we fine-tune it to our task using a regression loss (mean-squared error). BERTHA is then fine-tuned end-to-end following the same principals as in the BERT paper.

The two input sentences (system and human-generated captions) $\Gamma$ and $\mathrm{A}$  are tokenized using WordPieces \cite{wu2016google}. From the model's output, only the first symbol is used and the rest are discarded. A multilayer preceptor with a single output is attached to the $[CLS]$ symbol, the first output of BERT, to obtain the final score. 
$c$ is defined as $c \in \mathbb{R}^{H}$ where $H$ is the hidden side of the transformer blocks \cite{vaswani2017attention}.
We add a layer on top, with $w \in \mathbb{R}^{H}$ and $b\in \mathbb{R}$ as  weights and bias, to compute the final score $\gamma$:
\begin{equation}
    \gamma_{i,j,t} = \mathrm{softmax}(w^Tc_{i,j,t} +b),  
\end{equation}
which relates to the caption-reference pair for a single video $i \in [1,N']$, a specific reference $j\in [1,M']$  and a system $t\in [1,S]$  . 
%\subsection{Transfer learning}

%The pipeline of how both the model and the human judgment evaluated a system-generated caption and human reference is provided in Figure ~\ref{fig:bertha_diagram}.
%Both sentences are segmented with WordPiece~\cite{wu2016google}. Two extra symbols are employed (as in the original BERT~\cite{devlin2018bert}), one to mark the start of the captions ([CLS]) and another to mark the end of both sentences([SEP]). The model's first output, which is represented with the symbol $C$, is connected to a Fully Connected layer (FC) and softmax. This new layer will finally obtain the predicted score, $\gamma$, for the two input sentences. Human assessment of the two original sentences is carried out in advance. The human evaluates the adequacy of the caption and provides a score, which is averaged with other human scores for the same caption, producing $A$. This score is used to train BERTHA and also later  for evaluation of metric performance.

\section{Evaluation Methodology}
\label{results}

\subsection{Dataset}

We use datasets from the past TRECVid \cite{awad2020trecvid} benchmark video to text task (VTT) from 2016~\cite{1178722}, 2017~\cite{TRECvid17}, 2018~\cite{TRECvid18}, 2019~\cite{2020arXiv200909984A} and 2020~\cite{awad2021trecvid}. 
Approximately 2,000 videos are available each year from Vine and later also from Flickr and V3C2. 
The video covers multiple topics and multiple camera angles. The videos are short clips about 6 seconds long; some videos can be up to 10 seconds long in the last year. Each of these videos has been annotated with a caption by between two and five human captioners. 
The original tasks was divided in two subtasks. The first one, to generate a caption for each video and the second one to find and match human-generated captions with the correct video. 
We use the data to train and evaluate the model taken from captions produced by systems that submitted results for the first task.

\begin{table*}[htp]
    \centering
    \begin{tabular}{lrrrrrr}
\toprule
   & VTT-16     & VTT-17    & VTT-18 & VTT-19  & VTT-20 & Mean \\
 \midrule
    BERTHA-SA & 0.274 & 0.801 & 0.948 & \textbf{0.929} & 0.963& 0.910 \\
    BERTHA-MA & 0.720 & 0.706  & 0.878  &  0.837  & 0.863 & 0.821 \\
    BLEU-4 & 0.771 & 0.782 & 0.874 & 0.581 & 0.944 & 0.795\\
    Cider  & \textbf{0.811} & 0.818 & 0.961 & 0.810  &  \textbf{0.977} & 0.891 \\
    METEOR & 0.628 & \textbf{0.907} & \textbf{0.989}& 0.887 & 0.958 & 0.935 \\
    Rouge  & 0.295 & 0.555 & 0.926 & 0.588 & 0.919&  0.749 \\
    SPICE  & -0.278 & 0.103 & -0.447 & 0.498 & -0.110 & -0.033\\
\bottomrule
\end{tabular}

    \caption{Pearson correlation of the video captioning system-level metric scores with human assessment SA for TRECVid 2016 to 2020 participating systems; the metric is trained on captions from all other (non-test) years. The last column refers to the mean score of the metric across all years without taking into account VTT-16.}
    \label{results-low-sys}
\end{table*}
\subsection{Human Assessment}

The annotations from the human assessment were collected using Amazon Mechanical Turk\footnote{\url{http://www.mturk.com}}. The general idea of this process is that a set of humans will look at a system-generated caption and give a score between 0 to 100 on how similar is it to the original video. They also need to evaluate how fluent the caption is in English. Combining these two measures is the reference score that we use to train and test our system. To detect consistency in the same annotators, degraded and repeated sentences are used. This paper divides the annotations into two sets: single annotator (SA) and multiple annotators (MA).

The SA set is collected as in \cite{journalpone0202789} where only a single human annotates each pair of captions. The worker would see the video and then compare the sentence with what was seen. To validate that the annotators are reliable, a process was used where captions coming from both automatic systems and human annotators were used together. Some of the human captions are also degraded with random parts of other human captions. Thus, an inaccurate worker could be identified as they will overlook the captions produced by humans and not give them a high score. Equally, they will also skip the degraded captions and assign a poor score. The final score consists of a $z$-score computed as the standard deviation from the mean data point regarding its standard deviation and mean score. The filtering of the annotators is done before computing the $z$-score.

We detect some inconsistencies with the SA so we produce a new set, MA, which is similarly collected. However, instead of using a single annotator per system-references, a minimum of 15 annotations were used. The 15 annotations proved to be sufficient to obtain stability in the scoring at the segment-level as in \cite{grahametal:2015}. The final score is the standardisation of these annotations. 

In terms of size, the SA dataset has 56,088 human annotations that are not equal distributed across years, and MA dataset has 7,705 annotations equally distributed across each year, and MA data is more costly than collecting SA data. SA has a mean of 15 tokens per human reference caption, and nine tokens of the system generate captions in terms of length. MA has a mean of 14 tokens per human reference and eight tokens per system-generated captions.
%Also, notice that the equitable distribution of captions per year should not significantly impact as we are equally interested in low-quality captions and high qualities ones. 
\begin{table*}[htp]
    \centering
    \begin{tabular}{lrrrrrr}
\toprule
       & VTT-16 & VTT-17 & VTT-18 & VTT-19 & VTT-20  & Mean\\
        \midrule
BERTHA-SA & \textbf{0.857} & 0.744   & 0.936  & 0.859   & 0.888 & 0.856\\
BERTHA-MA & 0.743  & 0.715  & 0.884  & 0.882  & 0.863    & 0.836                  \\
BLEU-4 & 0.119  & 0.607  & 0.221  & 0.753  & 0.967      & 0.637                \\
CIDEr  & -0.001 & \textbf{0.848}  & 0.934  & \textbf{0.892} & 0.967   & 0.910                \\
METEOR & -0.449 & 0.792  & 0.942  & 0.880   & \textbf{0.988}   &   0.900            \\
Rouge  & -0.650  & 0.656  & \textbf{0.949}  & 0.811  & 0.914   &       0.642             \\
SPICE  & -0.356  & -0.135  & 0.079  & -0.240  & 0.146     &   -0.037         \\
\bottomrule
\end{tabular}

    \caption{Pearson correlation of the video captioning system-level metric scores with human assessment MA for TRECVid 2016 to 2020 participating systems; metric is trained on captions from all other (non-test) years. The last column refers to the mean score of the metric across all years without taking into account VTT-16.}
    \label{results-high-sys}
\end{table*}

One relevant statistic is how the scores from the human assessment are distributed in each dataset. Figure~\ref{fig:captions} shows the different distribution per year and dataset. SA has a mean score of -0.28, and MA has a mean score of -0.25; however, in the plot we can see that the deviations are very different. As both datasets include similar sentences we can deduce that same quality system generated sentences has a noticeable difference in the human score in each dataset.
Evaluation is performed at the system level and the caption level. However, the method is trained only at the caption level.

\section{Evaluating BERTHA in TRECVid}
In this section, the comparison of how BERTHA works in the two datasets is presented. We focus on two primary configurations: BERTHA-SA and BERTHA-MA. Each model is trained in all the years of one of the datasets, e.g.~BERTHA-SA is trained in the single annotator dataset. Each dataset is divided by each year to represent better the typical set-up of the TRECVid challenge, so we train in a set of years and keep one for testing. In each table, the column represents the year used as a test. In both models, the same year used for testing is discarded from training as some system-generated sentences are shared between the two datasets. 

Even if there are multiple references, they are not standard in number. In the early years of the challenge, only two human sentences per video were used. Also, all the videos and systems do not have the same number of human references. Because of this, the base experiment will consider each sentence independently, even if they come from the same video and the same system. In further experiments, we will discuss the use of multiple references.
\begin{table*}[htp]
    \centering
    % \begin{tabular}{lrrrrr}
% \toprule
%   & VTT-16     & VTT-17    & VTT-18 & VTT-19  & VTT-20  \\
%  \midrule
% BERTHA-SA & 8.8  & 7.7  & 24.3  & 22.6 & 6.9 \\
% BERTHA-MA & 5.0 &2.8 & 3.3 & 7.2 & 3.2\\
% BLEU-4 & 4.4  & 2.1  & 11.5  & 6.2 & 2.6 \\
% SentBLEU &  3.6  & 2.2  &  9.3  & 5.3 & 2.7 \\
% CIDEr  & 10.3  & 5.4  & 21.0   & 15.9 & 10.7 \\
% Meteor & 11.5  & 9.8  & 38.7  & 30.1  & 11.5 \\
% Rouge  & 8.6  & 2.7  & 13.0   & 5.1  & 8.2 \\
% SPICE  & 2.7  & 0.1  & 0.4   & 1.9  & 0.6 \\
% \bottomrule
% \end{tabular}

\begin{tabular}{lrrrrrr}
\toprule
   & VTT-16     & VTT-17    & VTT-18 & VTT-19  & VTT-20 & Mean  \\
 \midrule
BERTHA-SA & \textbf{0.081}  & 0.027  & \textbf{0.091}  & \textbf{0.075} & 0.069 & 0.069 \\
BERTHA-MA & 0.050 &0.028 & 0.033 & 0.072 & 0.032 & 0.041\\
BLEU-4 & 0.028  & 0.004  & 0.016  & 0.030 & 0.010 & 0.015\\
SentBLEU &  0.036  & 0.022  &  0.043  & 0.053 & 0.027 & 0.036\\
CIDEr  & 0.017  & \textbf{0.044} & 0.059   & 0.035 & 0.107 & 0.061 \\
METEOR & 0.027  & 0.034  & 0.087  & 0.064  & \textbf{0.115} & 0.075\\
Rouge  & 0.002  & 0.017  & 0.076   & 0.046  & 0.093 & 0.058 \\
SPICE  & 0.023  & 0.011  & 0.062   & 0.032  & 0.003 &  0.027\\
\bottomrule
\end{tabular}
    \caption{Pearson correlation of video captioning caption-level metric scores with human assessment SA for TRECVid 2016 to 2020 participating systems; the metric is trained on captions from all other (non-test) years. The last column refers to the mean score of the metric across all years without taking into account VTT-16.}
    \label{results-low-cap}
\end{table*}

\begin{table*}[htp]
    \centering
    % \begin{tabular}{lrrrrr}
% \toprule
%       & VTT-16 & VTT-17 & VTT-18 & VTT-19 & VTT-20\\
%         \midrule
% BERTHA-SA & 6.3 & 9.7 &    7.7 & 14.7 & 16.4\\
% BERTHA-MA & 4.0   & 6.6  & 13.2  & 22.5  & 24.7                      \\
% BLEU-4 & 1.7  & 45  & 0.7  & 15.5  & 11.2                      \\
% SentBLEU & 10.4  & 3.5 & 12.0  & -5.1  & 7.7                     \\
% CIDEr  & 4.9  & 8.3  & 11.4  & 22.2  & 20.8                      \\
% Meteor & 1.7  & 8.3  & 12.9  & 12.6  & 23.5                     \\
% Rouge  & 5.5  & 5.6  & 4.8  & 4.9  & 14.0                       \\
% SPICE  & 6.1  & 2.6  & 4.4  & -0.1  & -13.2                       \\
% \bottomrule
% \end{tabular}

\begin{tabular}{lrrrrrr}
\toprule
      & VTT-16 & VTT-17 & VTT-18 & VTT-19 & VTT-20 & Mean\\
        \midrule
BERTHA-SA & 0.063 & 0.060 &    0.077 & 0.147 & 0.164 & 0.112\\
BERTHA-MA & 0.042   & 0.066  & 0.058  & \textbf{0.225} & \textbf{0.247} & 0.149                    \\
BLEU-4 & 0.017  & 0.045  & 0.007  & 0.049  & 0.112  &  0.053                  \\
SentBLEU & 0.024  & 0.035 & 0.010  & -0.051  & 0.077    &   0.017               \\
CIDEr  & 0.049  &\textbf{0.083}  & 0.114  & 0.155  & 0.208   &  0.140                 \\
METEOR & 0.017  & \textbf{0.083} & \textbf{0.129}  & 0.222  & 0.235  &  0.167                 \\
Rouge  & \textbf{0.064}  & 0.056  & 0.085  & 0.158  & 0.209    &   0.127                \\
SPICE  & 0.013  & 0.001  & 0.012  & 0.018  & 0.001  &   0.008                  \\
\bottomrule
\end{tabular}
    \caption{Pearson correlation of video captioning caption-level metric scores with human assessment MA for TRECVid 2016 to 2020 participating systems; the metric is trained on captions from all other (non-test) years. The last column refers to the mean score of the metric across all years without taking into account VTT-16.}
    \label{results-high-cap}
\end{table*}

\subsection{System-level Evaluation}
%\subsubsection{Evaluation Process}
Concerning system-level evaluation, each metric takes as input the set of videos from a single past TRECVid-VTT task and one or more human-generated reference captions for that video, as well as the video captions produced by all systems participating in that task. 
For each participating system, the metric produces a single score corresponding to the average score of each sentence produced by this system.
Taking scores produced by the metric for all systems from a single TRECVid-VTT task, we calculate how well it correlates with human assessment of the same systems in terms of SA or MA scores described above.

To evaluate the system level, we use the following formula:
\begin{equation}
\rho(P_t,U_t),
\end{equation}
%
%where \[P=\bigg\{\frac{1}{N'M'}\sum_{i,j}^{N',M'}\gamma_{i,j,t} ,\quad t\in \mathbb{N}, t < S \bigg\} \;  \]	and \[U=\bigg\{\frac{1}{N'M'}\sum_{i,j}^{N',M'}  A_{i,j,t}, \quad t\in \mathbb{N}, t < S \bigg\} \; .\]
where \[P_t=\frac{1}{N'M'}\sum_{i,j=1}^{N',M'}\gamma_{i,j,t}    \]	and \[U_t=\frac{1}{N'M'}\sum_{i,j=1}^{N',M'}  A_{i,j,t}.\]
%\subsubsection{Results} 
Table \ref{results-low-sys} presents the results for the full SA dataset. Notice that VTT-16 was the first year where this evaluation system was implemented, explaining why BERTHA and the other metrics behave differently. VTT-16 is also the earliest year; assuming that there is an improvement in the models each year then it is also the year with the worst system-generated captions in general terms. 

In the case of the MA dataset, represented in Figure \ref{results-high-sys}, VTT-16 still is the most different set in terms of results across all metrics. In this case, one of the most stable metrics is CIDEr, which always gets one of the top-3 scores in all years. BERTHA gets a good score overall: it is the second or third best performing metrics in most cases. 

Finally, we perform a William significance test~\cite{graham:2015:EMNLP} at the system level. Considering $p< 0.05$ as the references significance threshold in the William test, CIDEr is the best scoring metric, but only obtaining a clear difference from the worst-performing ones. Differences between metrics are more pronounced in the MA dataset.

\begin{figure}[htp]
\centering
\begin{subfigure}[b]{0.25\textwidth}
  \centering
  \includegraphics[width=.9\linewidth]{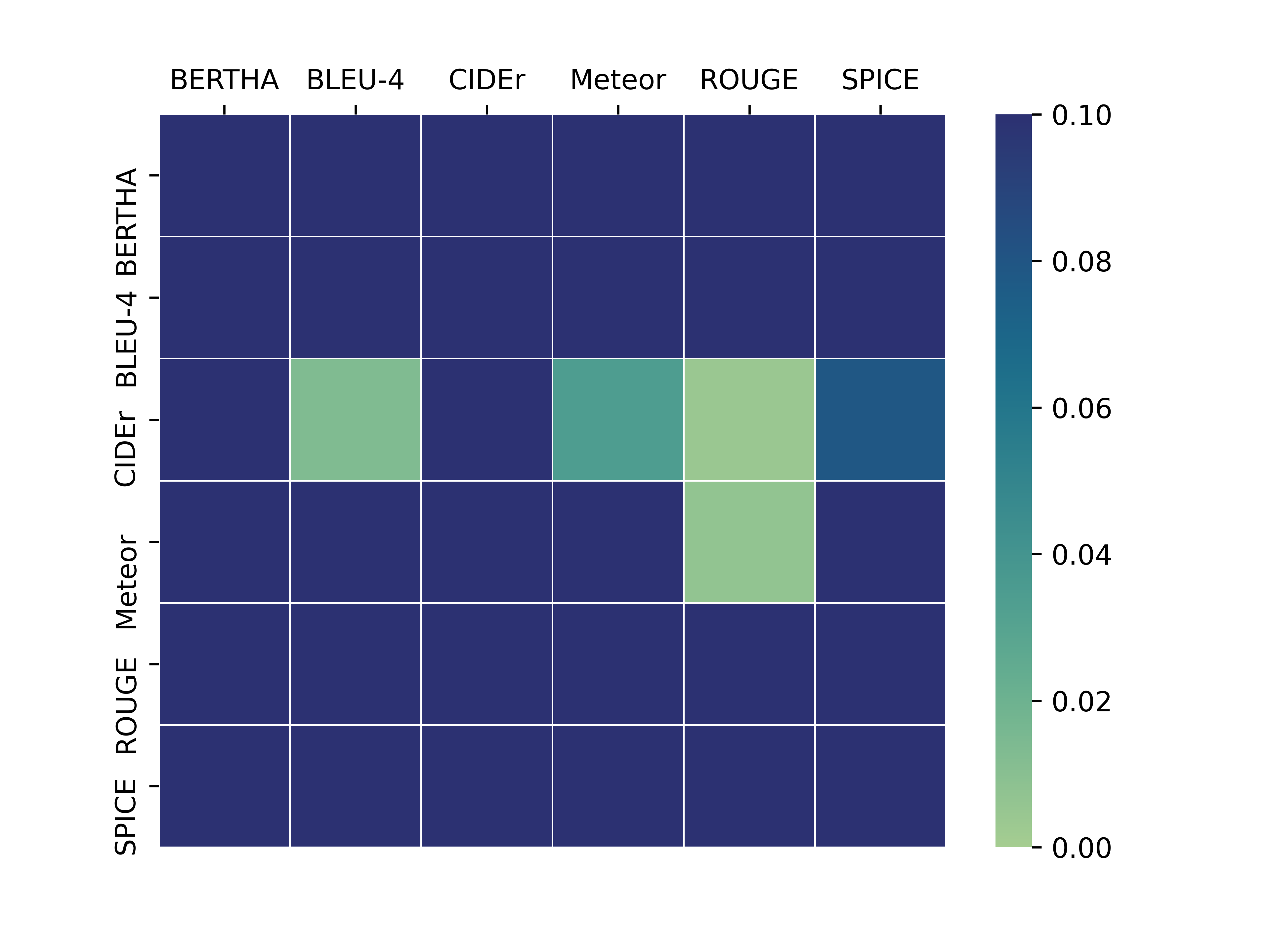}
  \caption{William significance test at the system-level metric scores for the SA dataset}
  \label{fig:WSAs}
\end{subfigure}%
\begin{subfigure}[b]{0.25\textwidth}
  \centering
  \includegraphics[width=.9\linewidth]{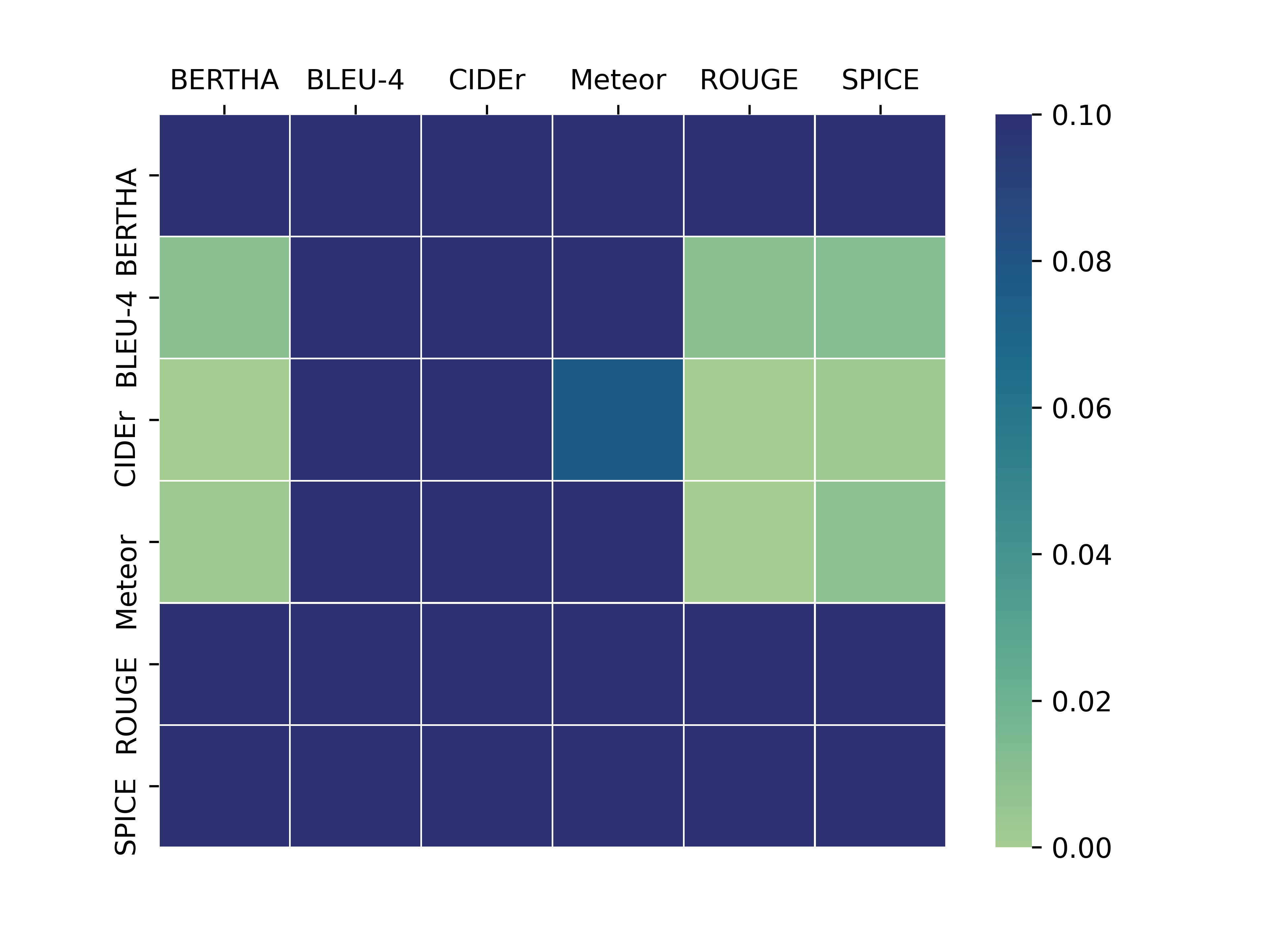}
  \caption{William significance test at the system-level metric scores for the MA dataset}
  \label{fig:wMAs}
\end{subfigure}
  \caption{William significance test of video captioning system-level metric scores with human assessment for TRECVid 2016 to 2020 participating systems. The $p$-value in a cell means the significance of the row vs the column in the absolute Pearson correlation.}
  \label{results-7williams}
\end{figure}

Overall the significance test shows no clear winning metric, justifying the use of multiple metrics to cover more situations. CIDEr obtains a better consistency in both datasets, but in terms of results, there are no significant differences compared to the other best-performing metrics.
 
In the case of BERTHA, we can see that VTT-16 performs worse than in the others years. From this, we can deduce that the tendency of VTT-16 is different from all the other SA years. However, the inclusion of this year as training seems not to affect the performance of BERTHA and it has similar behaviour to compared to the other metrics. We can argue that VTT-16 gives more examples of wrong captions, and the model aims to distinguish between good captions and incorrect captions. In the last two years, we can see an improvement in the overall scores that means that the captions were easier to catalogue, and there were better features to distinguish between good and bad sentences. 

\begin{table*}[htp]
    \centering
    \begin{tabular}{lrrrrrr}
\toprule
 & HUMAN-A & HUMAN-B & HUMAN-C & HUMAN-D & HUMAN-E & 5 Ref Captions \\
 \midrule
BERTHA-MA & \textbf{0.274} &\textbf{0.324}& \textbf{0.282}& \textbf{0.288}  & \textbf{0.288}& \textbf{0.236}\\
BLEU-4    & 0.089 & 0.082 & 0.042 & 0.074  & 0.100 & 0.057 \\
CIDEr     & 0.125 & 0.126 & 0.069  & 0.126 & 0.119 & 0.116 \\
METEOR    & 0.123 & 0.145 & 0.095 & 0.136 & 0.107 & 0.117  \\
\bottomrule
\end{tabular}
    \caption{Pearson correlation of video captioning caption-level metric scores with human assessment MA for the subset of TRECVid 2020 participating systems trained using a single human reference.}
    \label{results-5folds}
\end{table*}
\begin{table*}[htp]
    \centering
    
\begin{tabular}{lrrrrr}
\toprule
      & VTT-16 & VTT-17 & VTT-18 & VTT-19 & VTT-20 \\
      \midrule
BERTHA-MA & 0.037 & 0.061 & 0.117 & 0.158 & 0.217 \\
BLEU-4    & 0.002 & 0.042 & 0.047 & 0.002 & 0.000 \\
CIDEr     & 0.059 & 0.081 & 0.126 &0.170 & 0.222 \\
METEOR    & 0.014 & 0.077 & 0.136 & 0.218 & 0.242 \\
ROUGE & -0.088 & 0.036  & 0.048  & 0.116  & 0.166 \\
\bottomrule
\end{tabular}
    \caption{Pearson correlation of video captioning caption-level metric scores with human assessment MA for TRECVid 2016 to 2020 participating systems permuting the sentences evaluated.}
    \label{results-6permus}
\end{table*}

\subsection{Caption-level Evaluation}

In this case, the system is removed from the equation, and the metric is evaluated at the caption level. For each sentence produce by each system, we get a score. Taking all the scores for a year from a single TRECVid-VTT task, we calculate how well it correlates with human assessment of the same year.

To perform the caption level evaluation the $\rho(\Gamma_{N',S},A_{N',S})$ is used in the single references per caption scenario. Some videos have multiple references captions. Some traditional metrics, like CIDEr, have a mechanism to take into account these multiple sentences and obtain a better evaluation. BERTHA does not include an internal way to do it, so the mean of the predicted scores is used for the ranking:
\begin{equation}
\rho(P'_{i,t},U'_{i,t}),
\end{equation}
  where \[P'_{i,t}=\frac{1}{M'} \sum_{j=1}^{M'}\gamma_{i,j,t}\] and \[U'_{i,t}=\frac{1}{M'} \sum_{j=1}^{M'}  A_{i,j,t}.\]

Table \ref{results-low-cap} contains the results for the SA dataset at a caption-level. Overall the latest years are where the metric archive the best scores. CIDEr, METEOR, and BERTHA have similar correlations.

In the case of the MA dataset we can see a similar behaviour, as shown in Table \ref{results-high-cap}. BERTHA performs well and is the second or first metric with more similar performance to the human evaluation. Again the top performing metrics in all are BERTHA, CIDEr, and METEOR without strong discrepancies between them in the recent years.

\begin{figure}[htp]
\centering
\begin{subfigure}{.25\textwidth}
  \centering
  \includegraphics[width=.9\linewidth]{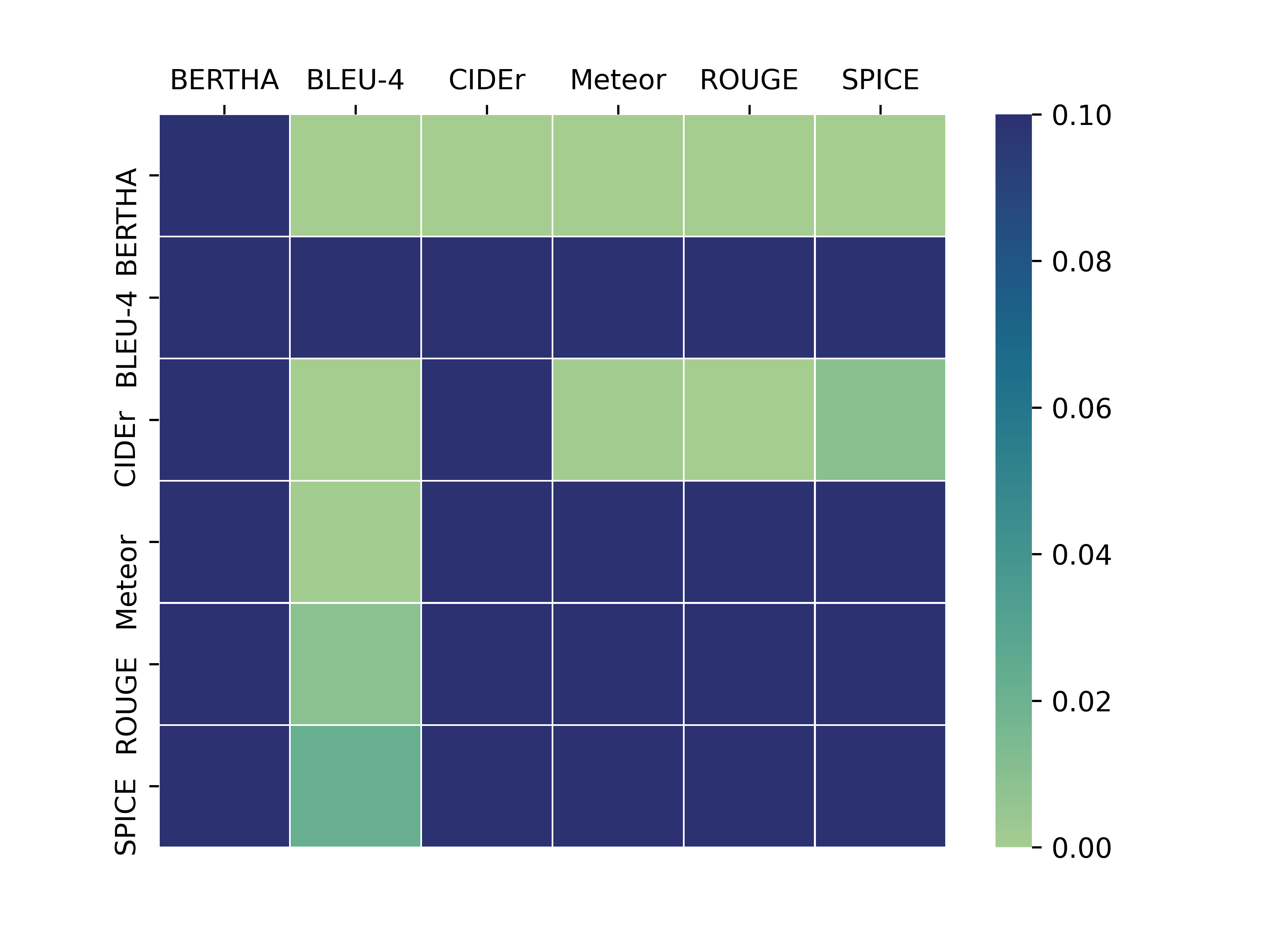}
  \caption{William significance test at the caption-level metric scores for the SA dataset.}
  \label{fig:WSAc}
\end{subfigure}%
\begin{subfigure}{.25\textwidth}
  \centering
  \includegraphics[width=.9\linewidth]{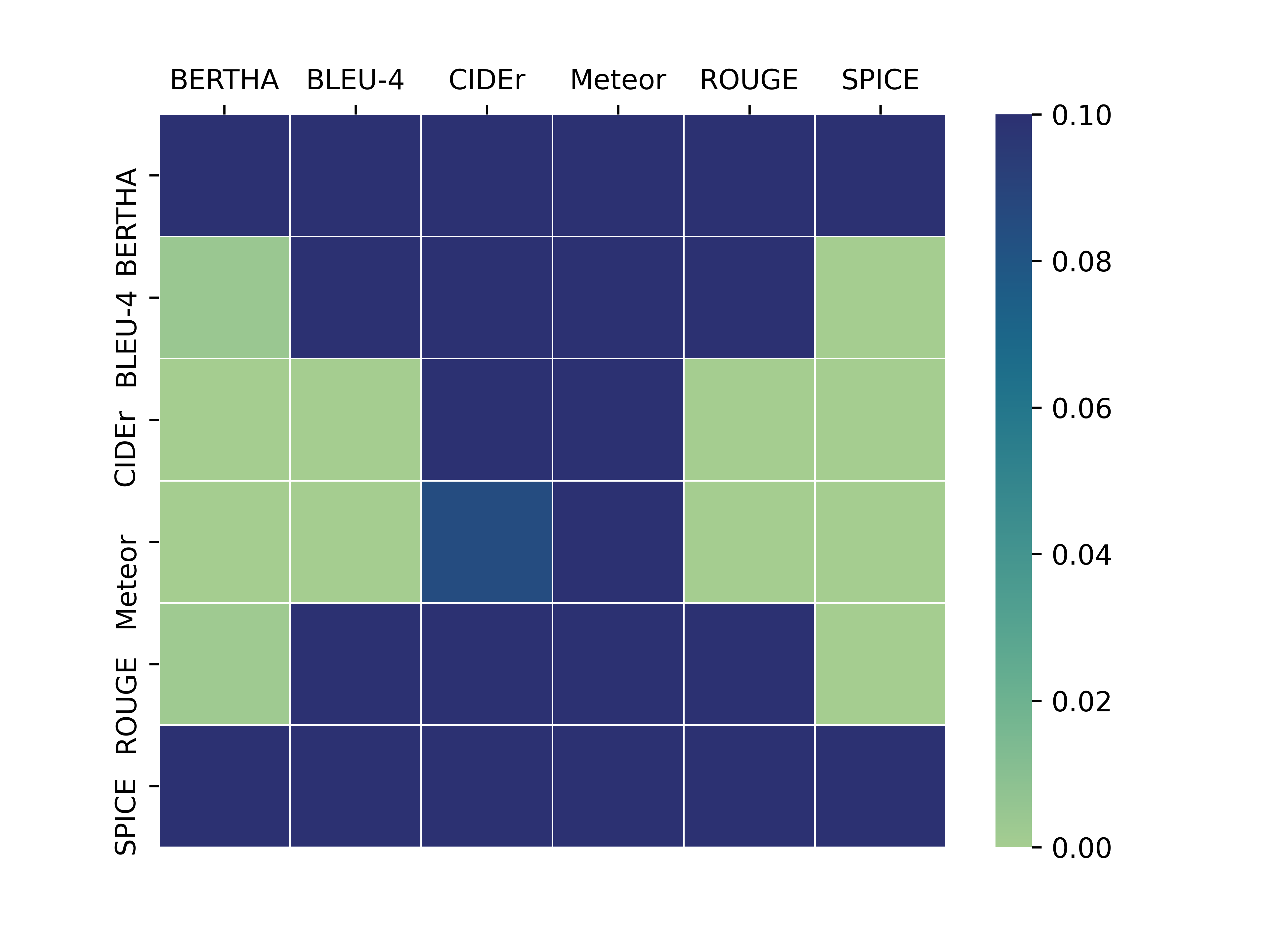}
  \caption{William significance test at the caption-level metric scores for the MA dataset.}
  \label{fig:wMAc}
\end{subfigure}
  \caption{William significance test of video captioning caption-level metric scores with human assessment for TRECVid 2016 to 2020 participating systems. The $p$-value in a cell means the significance of the row vs the column in the absolute Pearson correlation.}
  \label{results-8williams}
\end{figure}

Similar to the system-level evaluation, the William significance test was performed to evaluate the correlations. Figures \ref{results-8williams} show those results. BLEU-4 is the worst scoring metric in these scenarios. In MA, neither BERTHA nor SPICE 
achieve consistent results.

%\section{Discussion}

BERTHA outperform the other metrics in most of the test sets. In general, the MA dataset is the most complex dataset for BERTHA as it does not achieve good consistency, and the results vary depending on the year. 
%In the case of the MA data, VTT-16 become one of the worst-case for the rest of the metrics, dropping their score as some of the lowest levels. 

In general, we see a considerable drop in the correlation when comparing BERTHA train in MA and then the test in SA. On the contrary, the SA variant starts having slightly better results in the early years. However, in the last years of the task, it switches, and MA gets noticeable better results than other metrics. This means that the two datasets are not compatible, and the model can not be transferred between them in most scenarios. These cases are more relevant as the same sentence can exist in both datasets but have significantly different scores in each.
\begin{table*}[htp]
    \centering
    \resizebox{0.95\textwidth}{!}{%

\begin{tabular}{lrrrrrrrrrr} 
\toprule
       & \begin{tabular}[c]{@{}r@{}}SA-\\VTT-16\end{tabular} & \begin{tabular}[c]{@{}r@{}}SA-\\VTT-17\end{tabular} & \begin{tabular}[c]{@{}r@{}}SA-\\VTT-18\end{tabular} & \begin{tabular}[c]{@{}r@{}}SA-\\VTT-19\end{tabular} & \begin{tabular}[c]{@{}r@{}}SA-\\VTT-20\end{tabular} & \begin{tabular}[c]{@{}r@{}}MA-\\VTT-16\end{tabular} & \begin{tabular}[c]{@{}r@{}}MA-\\VTT-17\end{tabular} & \begin{tabular}[c]{@{}r@{}}MA-\\VTT-18\end{tabular} & \begin{tabular}[c]{@{}r@{}}MA-\\VTT-19\end{tabular} & \begin{tabular}[c]{@{}r@{}}MA-\\VTT-20\end{tabular}  \\ 
\midrule
BERTHA & 0.096                                                 & 0.02.3                                                 & 0.026                                                 & \textbf{0.087}                                        & 0.113                                                   & 0.020                                                 & \textbf{0.111}                                       & 0.091                                                 & 0.204                                                & 0.217                                               \\
BLEU-4 & 0.069                                                 & -0.001                                                & 0.023                                                 & 0.006                                                 & 0.043                                                                                                & -0.050                                                & -0.009                                                & 0.115                                                & 0.030                                                 & 0.110   \\
CIDEr  & 0.147                                                & 0.010                                                 & 0.033                                                 & 0.008                                                 & 0.131                                                & 0.028                                                 & 0.055                                                 & \textbf{0.238}                                      & 0.194                                                & 0.186                                                  \\
METEOR & \textbf{0.155}                                                & 0.009                                                 & 0.040                                                 & 0.038                                                 & 0.141                                                    & 0.048                                                 & 0.092                                                 & 0.151                                                & 0.234                                                & \textbf{0.264}                                               \\
ROUGE  & 0.121                                                & -0.011                                                & 0.026                                                 & 0.009                                                 & 0.115                                                 & -0.08                                                & 0.065                                                 & 0.083                                                 & 0.192                                                & 0.229                                                \\
Fusion     & 0.148                                       & \textbf{0.028}                                        & \textbf{0.041}                                        & 0.064       & \textbf{0.145}                                                & \textbf{0.069}                                        & 0.100                                                & 0.125                                                & \textbf{0.244}                                       & 0.233      \\
                                             
\bottomrule
\end{tabular}%
    }
    \caption{Pearson correlation of video captioning caption-level metric scores with human assessment SA and MA for TRECVid 2016 to 2020 participating systems using a combination of all the metrics (Fusion) to improve the final correlation.}
    \label{results-8psushing}
\end{table*}
\subsubsection{Using Multiple Human References}

In some years the dataset has multiple references: up to five per video. This allows comparing each system generated caption with different human references. Notice that there is no specific score for each human reference. It is shared between all the references to simplify the collection process. This is possible because the human references are closer to each other in terms of the human caption evaluation. 

Table \ref{results-5folds} shows the effect of using each of these references, named HUMAN-A, HUMAN-B, HUMAN-C, HUMAN-D, and HUMAN-E, and what happens when the five  are combined. The same humans did not annotated all the years, which is why we focus on a single year for this experiment. As a single year does not have sufficient data, the training strategy is to perform 5-folds on a subset of the data and show the mean. As BERTHA does not have any mechanism to use multiple references per video, the process to evaluate the five references per system generated caption is to compute the mean of the predicted scores before performing the correlation.  

In terms of results, BERTHA performs well  even in the small dataset. As BERTHA is a learned metric it obtains better results than in the multiple years set up. This shows that the metric is better suited when both the human evaluation and the system are more standardized. All the other metrics obtain worse numerical results than in the previous setup but they obtain a similar ranking compare with the previous setup.

In terms of analysing the humans, HUMAN-C produces the worst outcomes for all  metrics apart from BERTHA. This explains some differences between each human annotator, which in the end means that BERTHA does not perform as well in the five-captions scenario. We argue that this is the result of using the same score for five different references sentences for the same system generated caption. The system needs to learn to give the same exact score to five fairly different pair of sentences. The limited number of samples makes it difficult to generalize to this case.

\subsubsection{Evaluating Word Shuffle}
Table~\ref{results-6permus} shows what happens to the metrics when we train in a typical setting, but the sentences in evaluation have their words shuffled randomly. 

As expected, the metric most affected is BLEU, which gets a correlation near or equal to zero in some years. All the others, including BERTHA, are consistent with the previous experiments and only see minor variations, which means that even though they are slightly penalise by the shuffling of the words they are still able to reconstruct the meaning behind sentences. This demonstrates a certain robustness to miss-written sentences and that it is not only the grammar that is evaluated.

\subsubsection{Fusion of Metrics}

In this experiment a linear regression model is trained to find the best combination of metrics to predict the human judgment to produce a new fusion metric. All the years are divided in 80\% train and 20\% test. Then all the training sets are combined and a linear regression model is trained on this new dataset. Finally each year is tested separately. Table \ref{results-8psushing} shows the final correlation for all the years of both sub-datasets. All the metrics are in a range between 0 and 1, and no additional regularization was applied. The best fit linear regression coefficients are:
\\[1em]
{\small
\begin{tabular}{lllll}
  BERTHA  & BLEU-4 & CIDEr & METEOR & ROUGE \\
  \midrule
0.0525 & -0.1373 & 0.0315 & 0.2810 & -0.0779 
\end{tabular}}
\\[1em]

It can be observed that the fusion metric performs best in the SA dataset and performs as one of the top-3 in all the years of MA. This demonstrates that this fused metric is the most stable one. It correlates well with human judgment, making it a good reference for use in the evaluation of video captioning systems when a single metric is required.

From the regression weights, METEOR appears to be the base score for the final fusion metric while BLEU-4 is better used as a penalization when taking other metrics into account. A small weight indicates that a metric follows a similar behaviour as the main one. For instance, from Table \ref{results-high-cap} we can see that METEOR and BERTHA perform similarly; however, in the fused metric METEOR has a higher weight. 
%
%$0.05252609*BERTHA-0.13724702*BLEU-4+0.03151851*CIDEr+0.28098363*METEOR-0.07789074*ROUGE$.

\section{Conclusions}
\label{conclusions}
This paper proposed a new method to train an evaluation metric for video captioning. The technique learns how to compare system-generated captions with human references from human judgments. We study the robustness in a real challenge scenario. The dataset contains human assessment for system-generated captions.  BERTHA obtains a better correlation than the most commonly use metrics at the caption level as can be seen in Table \ref{results-low-cap} and \ref{results-high-cap}. From the Fusion analysis, we see that BERTHA can be used in a complementary way with the other metrics or as a standalone metric. BERTHA is purely a deep learning model, so it is easy to include in different architectures.

\section*{Acknowledgments}
This research was supported by the Irish Research Council Enterprise Partnership Scheme together with United Technologies Research Center Ireland and the Insight SFI Research Centre for Data Analytics supported by Science Foundation Ireland (SFI) under Grant Number SFI/12/RC/2289\_P2, co-funded by the European Regional Development Fund.

\section{References}
\bibliographystyle{lrec2022-bib}
\bibliography{mybibliography}
%
% \clearpage
% \section{Appendix: Extra things}
% \input{sec999_extra}
\end{document}